# Investigation by Driving Simulation of Tractor Overturning Accidents Caused by Steering Instability

Masahisa Watanabe [1*], Kenshi Sakai [2]

[1] Department of Food and Energy Systems Science, Graduate School of Bio-Applications and Systems Engineering, Tokyo University of Agriculture and Technology, Japan
[2] Division of Environmental and Agricultural Engineering, Institute of Agriculture, Tokyo University of Agriculture and Technology, Japan

*Corresponding author: s160479z@st.go.tuat.ac.jp

**ABSTRACT**
Overturning tractors are the leading cause of fatalities on farms. Steering instability contributes significantly to the tractor overturning. This study investigated tractor overturning accidents caused by the steering instability using a driving simulator. The general commercial driving simulator CarSim® (Mechanical Simulation Cooperation, MI, USA) was used. Tractor operations on steep passage slopes were simulated to mimic conditions present for a real accident case reported in Japan. Simulations were performed on roads with and without slopes. The tractor overturned only when on the road with the steep slope. The decrease in the vertical force on the front wheel caused the steering instability and the tractor to overturn. The steering instability caused understeer which prevents the operator from being able to control the tractor properly. Subsequently, the tractor overturned in the simulation. The tractor driving simulator was capable of reproducing the steering instability which can lead to the overturning accident.

## 1. INTRODUCTION

There are approximately 400 fatal farm accidents each year in Japan. Accidents involving agricultural tractors are a major contributor to farm fatalities. In 2016, 115 of the total 312 fatal farm accidents were tractor-related (Ministry of Agriculture, Fishery, and Forestry, 2018). More specifically, the tractor overturning is the leading cause of fatalities with 53 cases in 2016.

In Japan, small tractors specially designed for paddy fields are used in harsh environments such as rough farm roads, steep passage slopes, and narrow inclined side paths. This dangerous terrain can lead to a decrease in the vertical force on the front wheel. In some cases, this can result in separation of the front wheel from the underlying ground. This phenomenon causes vertical bouncing and lateral slippage of the tractor, both of which can lead to steering instability and overturning. The impact dynamics induced by the bouncing dramatically deteriorate tractor stability (Sakai, 1999; Sakai et al, 2000; Watanabe & Sakai, 2019a). If in addition to the bouncing slippage of the wheels occurs, the operator will not be able to maintain full control of the tractor. Consequently, the quality of the tractor posture dramatically decreases.

Several studies have contributed to the development of the tractor driving simulator and its application to farm safety and automation research (Gonzalez et al., 2017; Han et al., 2019; Watanabe & Sakai, 2019b). The tractor driving simulator is a strong platform for accident prevention research. The aim of the present paper is to apply the tractor driving simulator to investigation of overturning accidents induced by steering instability. A general driving simulator called CarSim® (Mechanical Simulation Cooperation, MI, USA) was used as a platform for the tractor driving simulator. Simulations of tractor operation on steep passage slopes were conducted. A real accident case reported in Japan was used as the basis for these simulations.

## 2. MATERIALS AND METHODS

The configuration of the tractor driving simulator is presented. CarSim® 2016 version was employed for the driving simulator. Vehicle and road configuration can be input by the user. Table 1 shows the tractor parameters used.

Table 1 Tractor parameter specification.

| Parameters | Value | Unit |
|---|---|---|
| Mass of tractor body | 788 | kg |
| Mass of wheels | 200 | kg |
| Pitch moment of inertia | 700 | kg m$^2$ |
| Distance between center of gravity of tractor body and front wheel | 0.7 | m |
| Distance between center of gravity of tractor body and rear wheel | 0.64 | m |

The road surface of the steep passage slope (on which the real accident case occurred) was recreated in the tractor driving simulator. According to the survey conducted by the Japanese Association of Rural Medicine, the tractor overturning accident happened on a steep passage slope of 19° gradient and 0.7 m in height (JARM, 2013). The tractor moved onto the passage slope from the farm field to the farm road and tried to turn right on the road to move into another farm field. However, the tractor was not able to turn and fell from the road. The road surface and scenario were configured in the driving simulator. To investigate the influence of the steep slope on the steering instability, two different types of the road surface were compared. Namely, with slope and without slope. Figure 1a and b shows the road surface with slope and without slope, respectively.

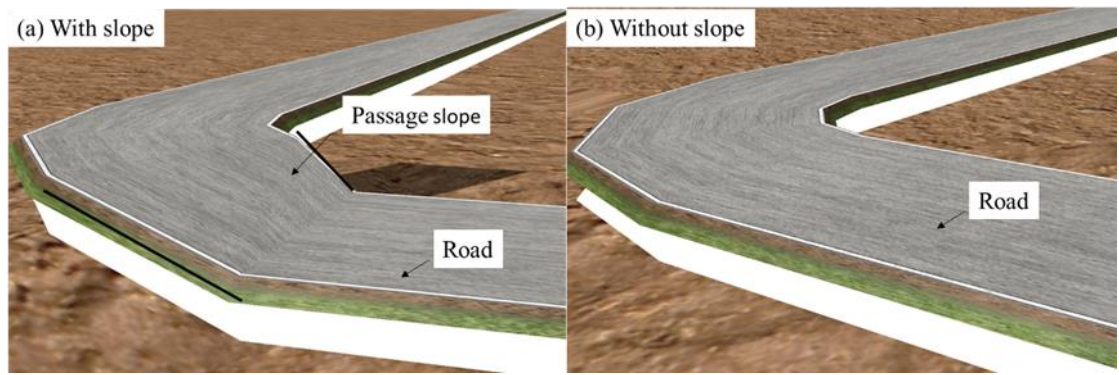

Figure 1 (a) Road surface with a slope; (b) Road surface without a slope.

Figure 2 shows the road profile of the slope.

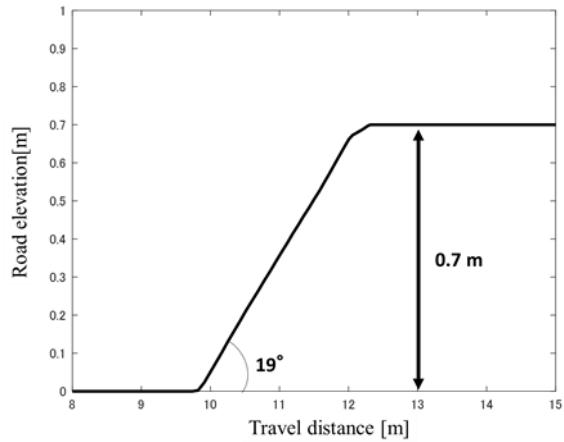

Figure 2 Profile of the slope. The gradient is 19° and the height is 0.7 m.

## 3. RESULTS AND DISCUSSION

The velocity of the tractor was set to 4.3 m/s in the simulation. The tractor was ran on the road with slope and without slope. Figure 3 shows the tractor trajectories on the road in each simulation.

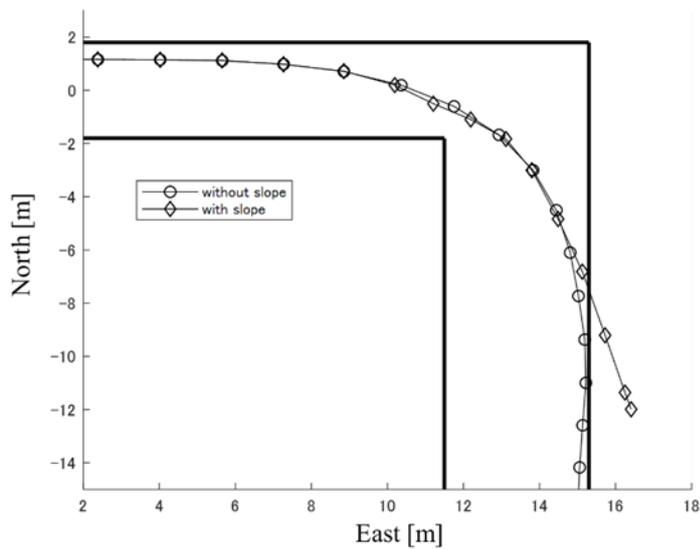

Figure 3 Tractor trajectories of each simulation.

The tractor remained in contact with the road during the whole simulation when the tractor ran on the road without slope. In contrast, the tractor ran off the road and then overturned when the tractor ran on the road with slope. To visualize the numerical results, Figure 4 and 5 show the animation of the driving simulation for the simulation without slope and with slope, respectively.

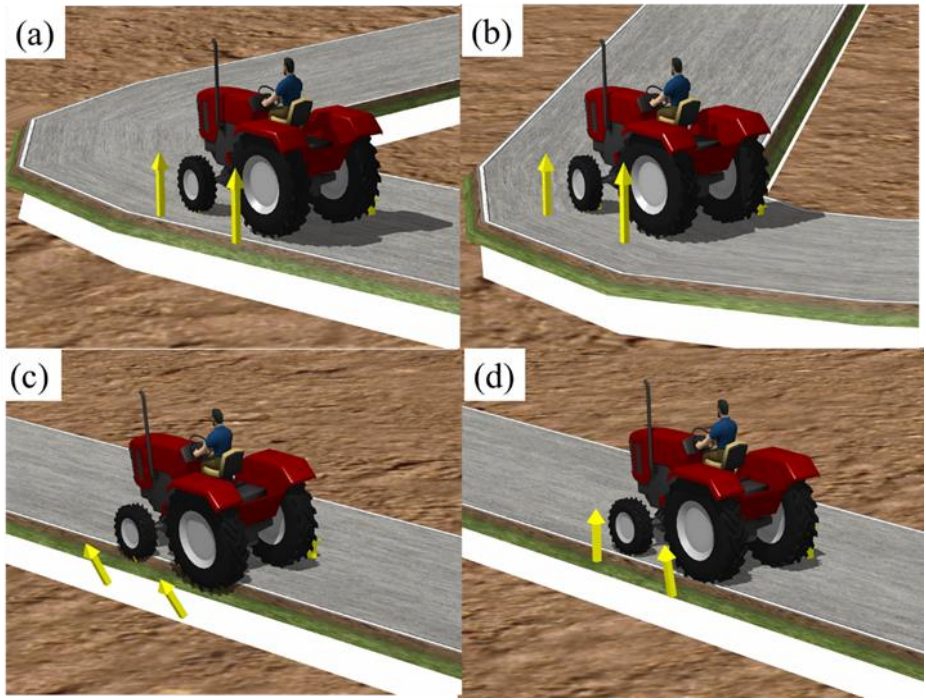

Figure 4 Animation of the tractor operation on the road without slope. (a) Tractor moved onto the corner; (b) Tractor ran on the corner; (c) Tractor was on the edge of the road; (d) Tractor continued to run without overturning.

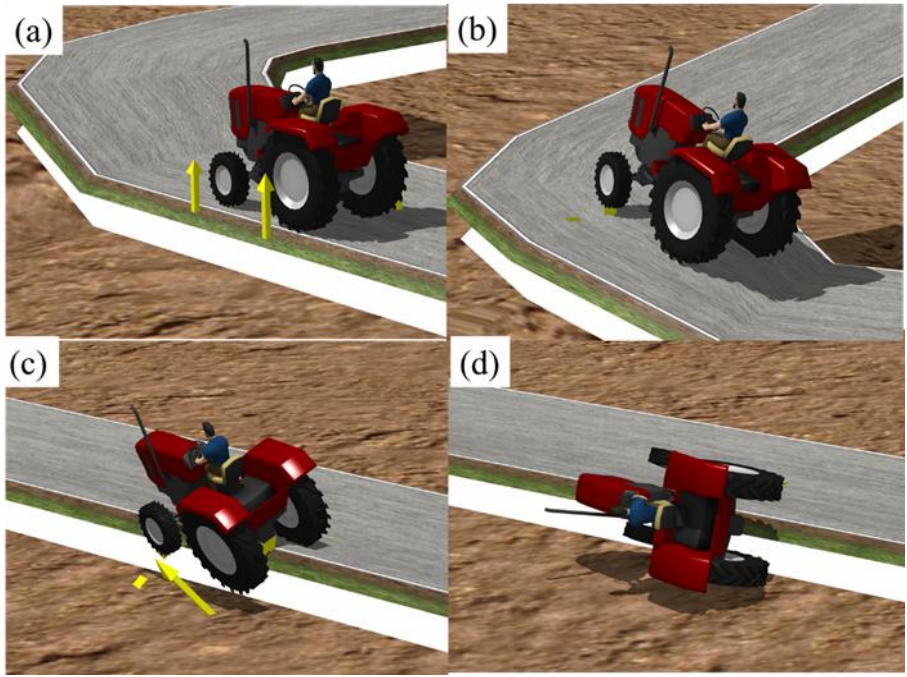

Figure 5 Animation of the tractor operation on the road with slope. (a) Tractor moved onto the slope; (b) Tractor ran on the slope; (c) The wheels went off the road; (d) Tractor overturning occurred.

Figure 6a and b show the vertical force on the front wheel and the cornering force on the front wheel, and the road elevation and the steering angle of the operator, respectively.

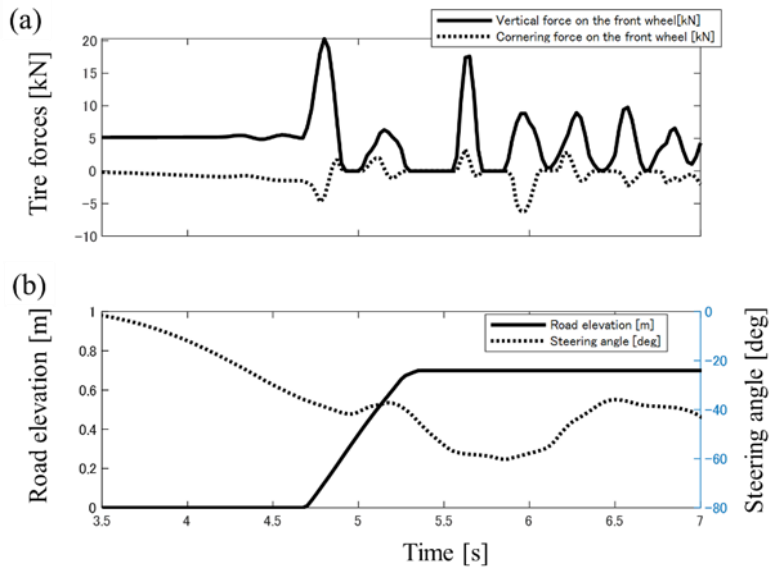

Figure 6 (a) The vertical force and the cornering force on the front wheel; (b) Road elevation and the steering angle of the operator.

When the front wheel of the tractor moved onto the slope, vibrations were induced and the vertical force on the front wheel decreased to zero as the road elevation increased. This caused the cornering force to be zero. Consequently, the operator cannot maintain control of the tractor and steering instability occurred. The steering instability caused understeer of the tractor and overturning. The results indicated that the tractor driving simulator could reproduce the steering instability which can lead to overturning.

## 4. CONCLUSION
The simulations of the tractor operations on the steep passage slope were conducted using the tractor driving simulator. Tractor overturning occurred in the simulation due to the steering instability. Future research will investigate how to avoid overturning by steering and develop accident prevention control for the overturning.

**ACKNOWLEDGMENT**
We thank Prof. Shrini Upadhyaya and Prof. Heinz Bernhardt for their kind support. This work was supported by JSPS Grant-in-Aid No. 19J11183 and 19H00959. We thank Edanz Group (www.edanzediting.com/ac) for editing a draft of this manuscript.